%% file: main.tex
\documentclass{article} 
\usepackage{iclr2021_conference,times}

\input{math_commands.tex}

\usepackage{hyperref}
\usepackage{url}
\usepackage{graphicx}
\usepackage{subcaption}
\usepackage{xspace}
\usepackage{microtype}

\newcommand*\rotate{\rotatebox{-90}}
\def\bphi{{\boldsymbol\phi}}
\def\x{\mathbf x}
\def\y{\mathbf y}

\def\w{\mathbf w}

\newcommand{\nomad}{{\sf NOMAD}\xspace}

\title{Efficient Training Under Limited Resources}

\author{Mahdi Zolnouri \\
Huawei Noah's Ark Lab\\
Montr\'eal, QC, Canada\\
\texttt{mahdi.zolnouri@huawei.com}
\And 
Dounia Lakhmiri\\
GERAD and Polytechnique Montr\'eal\\
Montr\'eal, QC, Canada\\
\texttt{dounia.lakhmiri@polymtl.ca}
\And  
Christophe Tribes\\
GERAD and Polytechnique Montr\'eal\\
Montr\'eal, QC, Canada\\
\texttt{christophe.tribes@polymtl.ca}
\And 
Eyyüb Sari \\
Huawei Noah's Ark Lab\\
Montr\'eal, QC, Canada\\
\texttt{eyyub.sari@huawei.com}
\And 
S\'ebastien Le~Digabel\\
GERAD and Polytechnique Montr\'eal\\
Montr\'eal, QC, Canada\\
\texttt{sebastien.le.digabel@gerad.ca}
}

\iclrfinalcopy 
\begin{document}

\maketitle
\renewcommand{\arraystretch}{1.3}
\begin{abstract}
Training time budget and size of the dataset are among the factors affecting the performance of a Deep Neural Network (DNN). This paper shows that Neural Architecture Search (NAS), Hyper Parameters Optimization (HPO), and Data Augmentation help DNNs perform much better while these two factors are limited. However, searching for an optimal architecture and the best hyperparameter values besides a good combination of data augmentation techniques under low resources requires many experiments. We present our approach to achieving such a goal in three steps: reducing training epoch time by compressing the model while maintaining the performance compared to the original model, preventing model overfitting when the dataset is small, and performing the hyperparameter tuning. We used \nomad, which is a blackbox optimization software based on a derivative-free algorithm to do NAS and HPO. Our work achieved an accuracy of 86.0\% on a tiny subset of Mini-ImageNet~\citep{vinyals2016matching} at the ICLR 2021 Hardware Aware Efficient Training (HAET) Challenge and won second place in the competition. The competition results can be found at \href{https://haet2021.github.io/challenge}{haet2021.github.io/challenge} and our source code can be found at \href{https://github.com/DouniaLakhmiri/ICLR_HAET2021}{github.com/DouniaLakhmiri/ICLR\_HAET2021}.

\end{abstract}

\section{Introduction}
The compression of DNNs targets mainly the inference side, but recent works aim to squeeze the training process. However, it remains challenging. ICLR 2021 HAET Challenge is an annual competition that evaluates the performance of classification models given a limited training time and data. The time budget for training is 10 minutes on an NVIDIA GPU V100 with 32 GB memory, running with an Intel(R) Xeon(R) Gold 6230 CPU @ 2.10GHz processor, with 12 GB of RAM. The dataset is a tiny unknown dataset of 10 classes containing 5K images for training and 1K images for testing during the development phase. The inputs to the model are 32$\times$32 RGB images like CIFAR-10. A subset of Mini-ImageNet with 80$\times$80 images is used to evaluate the performance of candidate models during the evaluation phase. Applicants are allowed to use their own optimizer, training loop, and data augmentation process.
Introducing such limitations to the training process creates two main challenges. First, most classifiers need much more than 10 minutes to converge due to a large number of parameters. The general trend in most of the state of the arts of DNNs is to go deeper and wider and increase the model's capacity~\citep{simonyan2014very,szegedy2016rethinking, szegedy2017inception,he2016deep}.
A larger network requires more time and computational resources to achieve its best accuracy. In this paper, we used a compressed model by applying Neural Architecture Search to reduce the computation operations of the model.
Second,  a small dataset leads the model to overfit. In this case, the model learns too well the training images and performs accurately on the training dataset but poorly on real-world images. Since Data Augmentation (DA) is an effective technique to generate additional data for training~\citep{taylor2017improving}, we added a combination of several DA policies to our pipeline to overcome the overfitting problem.    
\section{Related Work}
Since AlexNet~\citep{krizhevsky2012imagenet} won the 2012 ImageNet ILSVRC-2012 competition by stacking the convolution layers, the DNNs became more accurate thanks to increasing the number of parameters and hidden layers. In 2014, both VGGNet~\citep{simonyan2014very} with 19 layers and GoogleNet~\citep{szegedy2015going} with 22 layers achieved the lowest error rate in localization and classification tasks of the ILSVRC competition, respectively. In 2015, ResNet, the winner of ILSVRC, proposed the skip connection to overcome the vanishing gradient issue in DNNs and go even deeper by stacking layers up to 152. SENet~\citep{hu2018squeeze} won the ImageNet competition in 2017. It has 145M parameters almost twenty times bigger than GoogleNet with 6.8M. Recently, NFNet-F4+~\citep{brock2021high} achieved 89.2\% of top 1 accuracy and introduced a new state-of-the-art for ImageNet dataset using 527M parameters.
On the other hand, some networks tried to present a good trade-off between the model's size and accuracy.
For example, MobileNet~\citep{howard2017mobilenets} used three multipliers for depth, width, and the resolution of the model. It showed that users could reduce the model's size with the cost of losing accuracy. EfficientNet~\citep{tan2019efficientnet} used NAS to design a new baseline model. It introduced a new scaling method to build a new family of networks that achieved the state-of-the-art.
Mainly the focus is on inference side in these works and the proposed models are trained with no limitation in terms of data and resources during training process.

\section{Proposed Approach}\label{sec:proposed-approach}
\subsection{Simulating The Challenge Evaluation}\label{sec:simulation_s}
As an initial step, we started with implementing a training pipeline. We based on a provided code from the challenge website, and we added new modules such as data preprocessing, model building, and trainer. In the preprocessing module, we added a data sampler to build a dataset requested by the competition: 5k images for the train set and 1k images for the validation set. We selected CIFAR-10 as a proxy dataset because it is the most common dataset with ten classes in classification. Its size allowed us to build various subsets by shuffling the selected image indices before subsampling the data and helped our candidate model generalize better. Model building module provides a designed model as we previewed a known model's NAS variant. The trainer module runs the training loop for precisely ten minutes to simulate the competition environment. 

\subsection{Searching For A Baseline Network}\label{sec:baseline_search_s}
The larger the search space, the longer and more expensive NAS will be. Efficient use of resources was a top priority for the challenge. Therefore, instead of starting an architecture design from scratch, the second step of our approach was finding a baseline network candidate from which we would start the search. We came up with a list of well-known classifiers, mainly the winner of the ImageNet ILSVRC challenge. We adapted the implementation of most of these classifiers from~\cite{liukuang2021}. Then, we trained each classifier ten times with distinct subsets drawn from CIFAR-10. Training networks multiple times was important to smooth the variance effect induced by the small training sets. We trained all classifiers using our pipeline with 10 minutes training time budget and the same dataset. Other hyperparameters were similar to the respective paper of each model. We finally selected the top two models with the highest average accuracy. Table~\ref{tab:search_baseline_t} shows that SENet with~77.0\% and ResNet-18 with~75.5\% of validation accuracy are on top of the list.

\subsection{Neural Architecture Search}\label{sec:nas_s}
In order to reduce the training epoch time, we performed Neural Architecture Search on SENet and ResNet-18 by using Formulation~\ref{eq:nas_constrained}. The goal of this third step was to find a smaller variant of the model that can perform higher or equal to the baseline model. We used the \nomad~\citep{Le09b} blackbox optimization software which implements the MADS algorithm~\citep{AuDe2006}.
\nomad allows optimizing an objective function under certain constraints without requiring knowledge about the internals of the function. We defined $g(\bphi)$ as a function that denotes the resource consumption, an explicit function such as the number of Multiply-Accumulate (MAC) operations. We also constrained the model's performance to be higher or equal to the baseline one. We followed~\cite{howard2017mobilenets} to scale the baseline model by defining three multipliers for three dimensions of the model: depth, width, and input resolution. The depth multiplier determines the number of blocks for a model. The role of the width multiplier is to thin a network by reducing the number of input and output channels of each layer uniformly. Input resolution multiplier scales the input image and subsequently the internal representation of every layer. For each NAS trial, we return the number of MAC calculated based on the new dimensions of the model and the accuracy of the model to \nomad.

\begin{align}
\min_{\bphi \in \Phi} \quad        &g(\bphi)~~ \mathrm{subject~to~} ~~ f(\w,\bphi\mid\x,\y ) \geq f(\w_0,\bphi_0\mid \x,\y) \label{eq:nas_constrained}
\end{align}

Where $\w$ is the neural networks weights, $\Phi$ denotes the hyperparameter space of the neural network, $\bphi \in \Phi$ represents a vector of values for depth, width, and resolution that define the network architecture, and $(\x,\y)$ the data features and labels. We note $f(\bphi, \w\mid\x,\y)$ the validation accuracy of the network after training; And $g(\bphi)$ the function that reflects the number of MAC operations.

\subsection{Data Augmentation Method}\label{sec:da_s}
Given the small size of the dataset, we observed that our candidate model is prone to overfitting. Since the dataset is tiny, the model does not gain enough performance during training by using the basic techniques of image augmentation.
We used the AutoAugment~\citep{cubuk2018autoaugment} approach, which is an automatic way of selecting optimal data augmentation policies. Also, we added the CUTOUT~\citep{devries2017cutout} technique which consists of masking out random sections of input images during training.  

\subsection{Hyper Parameter Optimization}\label{sec:hpo_s}
Finally, we perform the HPO on our candidate model to find the best value of the hyperparameters by using \nomad. Formulation~\ref{eq:nas_unconstrained} shows the objective function that we proposed.

\begin{align}
\max_{\lambda \in \Lambda}\quad & f(\lambda, \w \mid \x, \y)   \label{eq:nas_unconstrained}
\end{align}

Where $\lambda$ is a vector of values from hyperparameter space of $\Lambda$. $f(\lambda, \w \mid \x, \y)$ denotes the validation accuracy of the neural network. We selected three standard hyperparameters known to strongly influence validation set performance: initial learning rate, weight decay, and the optimizer type. Since the dataset is tiny, we chose to tune the batch size as well. As the range of values for learning rate is different from an optimizer to another, we adapted the learning rate based on an expected value of learning rate for each optimizer during training. For example, $0.1$ is usually used for SGD, but $0.001$ is used for ADAM. In the end, the search space of the HPO experiment was as follows:
\begin{itemize}
 \item Learning rate of uniform type and in [0.6;1E-3]
 \item Weight decay of discrete type and in \{0, 0.00005,  0.0005, 0.005, 0.05, 0.5\}
 \item Optimizer of discrete type and in \{``Adadelta"(1) , ``Adagrad" (0.01) , ``SGD" (0.1), ``Adam" (0.01), ``Adamw" (0.01),  ``Adamax" (0.002),  ``ASGD"\}
 \item Batch size of discrete type and in \{128, 256, 512\}
\end{itemize}

The HPO increased the accuracy of our model by 1.1\%.
\section{Experiments and Results}\label{sec:experiments-and-results}
In this section, we conduct experiments to investigate the effectiveness of our approach across a range of model architecture and datasets.

\subsection{CIFAR-10}
CIFAR-10 is a collection of 60K color images of 32$\times$32, divided as follows: 50K for train set and 10K for validation set. It has ten classes. Since it is so close to the description of the competition evaluation dataset and its size is big enough to have several distinct subsamples, we decided to use it during the development phase as a proxy dataset. In all our experiments, the size of the dataset is as follows: the train set is 5K, and the validation set is 1K.
Table~\ref{tab:search_baseline_t} shows the intermediate results of our search to find a baseline model. All the classifiers are trained for ten minutes and on a subset of CIFAR-10. SENet-18 with 77.0\% and ResNet-18 with~75.5\% achieved better accuracy than other classifiers.

\begin{table}[h!]
    \centering
    \footnotesize
    \setlength\tabcolsep{2pt} 
    \begin{tabular}{l  c c c c c c c c c c c c c c c c c c} \\
      \hline
        &  \rotate{SENet-18} & \rotate{ResNet-18} & \rotate{DenseNet121} & \rotate{GoogleNet} & \rotate{MobileNet V2} & \rotate{EfficientNet B0} & \rotate{ShuffleNet V2} & \rotate{RegNetX\_200MF} & \rotate{ResNet32} & \rotate{SimpleDLA} & \rotate{ResNet50} & \rotate{VGG19} & \rotate{MobleNet V1} & \rotate{ResNeXt29~(2x64d)~~} & \rotate{ShuffleNet G2} & \rotate{ResNet110} & \rotate{PreActResNet18} & \rotate{DPN92}  \\
     \hline\\
     {Accuracy} & \textbf{77.0} & 75.5 & 74.6 & 74.2 & 74.0 & 70.4 & 70.0 & 69.8 & 69.6 & 68.7 & 67.6 & 67.2 & 64.8 & 64.7 & 62.0 & 58.3 & 52.7 & 41.9\\
     \hline
\end{tabular}
    \caption{Searching for a baseline network. Final validation accuracy (\%): mean over ten runs. }
    \label{tab:search_baseline_t}
\end{table}

By applying NAS on SENet-18 and ResNet-18, we were able to find a smaller variant for each model that performs similarly to its baseline model. The NOMAD-NAS-SENet-18 has the same depth and input resolution as SENet-18, but the width of the network is 67\% of the original one. The NOMAD-NAS-ResNet-18 has the same depth as ResNet-18, but its width and input resolution are 73\% and 118\% of the original ones, respectively. Although the input resolution is 18\% bigger than its original model, the NOMAD-NAS-ResNet-18 has 22\% less MAC operations and 47\% fewer parameters than ResNet-18. We observed that both candidate models are prone to overfitting. We improved our data transformation list by adding AutoAugment and CUTOUT techniques. Table~\ref{tab:top_2_after_nas_autoaugment_t} shows the performance of SENet-18 and ResNet-18 after applying NAS and DA. After this step, we continued on hyperparameters optimization with NOMAD-NAS-SENet-18. Table~\ref{tab:hpo_t} shows the validation accuracy of our candidate model, NOMAD-NAS-SENet-18 after applying HPO. We used \nomad to find the best hyperparameter values of the model. To this end, we performed a blackbox optimization by using MADS algorithm over multiple parameters of our final candidate model. As described in~\ref{sec:hpo_s}, we came up with a set of tuned values for the hyperparameters: learning rate = 0.042, weight decay = 0.005, optimizer = SGD, batch size = 512. We performed several tests to ensure these best values can achieve optimal accuracy. We observed an improvement of validation accuracy from~86.7\% to~87.8\%.

\begin{table}[h!]
    \centering
    \caption{Applying NAS and DA on SENet-18 and ResNet-18. Final validation accuracy: mean over ten runs.}
    \label{tab:top_2_after_nas_autoaugment_t}    
    \input{tables/top_2_after_nas_autoaugment}
\end{table}

\begin{table}[h!]
    \centering
    \caption{Applying HPO on NOMAD-NAS-SENet-18. Final validation accuracy: mean over ten runs.}
    \label{tab:hpo_t}    
    \input{tables/hpo_cifar10}
\end{table}

\subsection{Mini-ImageNet}
After the competition deadline, the ICLR HAET Challenge committee announced the results and the name of the evaluation dataset. Table~\ref{tab:final_results_mini_imagenet_t} shows the final result of our model on a subset of Mini-ImageNet. There is a considerable gap of 1.8\% between our evaluation on CIFAR-10 and committee evaluation on Mini-ImageNet. Some other potential differences, such as hardware and training loop, can justify this gap. However, the main factor affecting our model's performance is an extra operation in data transformation that wasn't included in the challenge description. The images in Mini-ImageNet are 80$\times$80. The committee added a resize operation to the data transformation list in order to convert images to the initial announcement of 32$\times$32. We used Cosine Annealing for the learning rate decay scheduler with the max epoch of 240 for ten minutes. This extra data transformation is costly and delayed our training process for several epochs, consequently reducing our model's performance. We found the results fair because this extra cost affects all other submissions. 

\begin{table}[h!]
    \centering
    \caption{NOMAD-NAS-SENet-18 on Mini-ImageNet. Final validation accuracy: mean over five runs. Dataset is a subset of Mini-ImageNet : train set of 5K images and validation set of 1K images}
    \label{tab:final_results_mini_imagenet_t}    
    \input{tables/final_results_mini_imagenet}
\end{table}




\section{Conclusion}
ICLR 2021 HAET Challenge focuses on training a neural network under low resources, more precisely limited data and training time. Most research done in the model compression field tried to optimize the inference part of DNNs. In order to train a model efficiently, we proposed an approach that finds a classifier that performs accurately on a tiny dataset and a limited training time budget. We used \nomad, which implements a derivative-free optimization algorithm for both NAS and HPO. Also, we used data augmentation techniques to improve the performance of the proposed model. Our model, NOMAD-NAS-SENet-18, achieved~86\% on a subset of Mini-ImageNet under 10 minutes and won second prize in the competition.
\newpage


\bibliography{biblio}
\bibliographystyle{iclr2021_conference}


\end{document}

%% file: math_commands.tex
\usepackage{amsmath,amsfonts,bm}

\def\eqref#1{equation~\ref{#1}}

\def\1{\bm{1}}

\DeclareMathAlphabet{\mathsfit}{\encodingdefault}{\sfdefault}{m}{sl}
\SetMathAlphabet{\mathsfit}{bold}{\encodingdefault}{\sfdefault}{bx}{n}



%% file: tables/top_2_after_nas_autoaugment.tex
\begin{tabular}{ l | c }
    \hline
    Model &  Validation Accuracy (\%) \\
    \hline
    SENet-18 & \textbf{86.7} \\
    ResNet-18 & 84.8 \\
    \hline
\end{tabular}



%% file: tables/hpo_cifar10.tex
\begin{tabular}{ l | c }
    \hline
    Model &  Validation Accuracy (\%) \\
    \hline
    NOMAD-NAS-SENet-18 & \textbf{87.8} \\
    \hline
\end{tabular}

%% file: tables/final_results_mini_imagenet.tex
\begin{tabular}{ l | c }
    \hline
    Model &  Validation Accuracy (\%) \\
    \hline
    NOMAD-NAS-SENet-18 & 86.0 \\
    \hline
\end{tabular}

%% file: main.bbl
\begin{thebibliography}{17}
\providecommand{\natexlab}[1]{#1}
\providecommand{\url}[1]{\texttt{#1}}
\expandafter\ifx\csname urlstyle\endcsname\relax
  \providecommand{\doi}[1]{doi: #1}\else
  \providecommand{\doi}{doi: \begingroup \urlstyle{rm}\Url}\fi

\bibitem[Audet \& {Dennis, Jr.}(2006)Audet and {Dennis, Jr.}]{AuDe2006}
C.~Audet and J.E. {Dennis, Jr.}
\newblock {Mesh Adaptive Direct Search Algorithms for Constrained
  Optimization}.
\newblock \emph{SIAM Journal on Optimization}, 17\penalty0 (1):\penalty0
  188--217, 2006.
\newblock \doi{10.1137/040603371}.

\bibitem[Brock et~al.(2021)Brock, De, Smith, and Simonyan]{brock2021high}
Andrew Brock, Soham De, Samuel~L Smith, and Karen Simonyan.
\newblock High-performance large-scale image recognition without normalization.
\newblock \emph{arXiv preprint arXiv:2102.06171}, 2021.

\bibitem[Cubuk et~al.(2018)Cubuk, Zoph, Mane, Vasudevan, and
  Le]{cubuk2018autoaugment}
Ekin~D Cubuk, Barret Zoph, Dandelion Mane, Vijay Vasudevan, and Quoc~V Le.
\newblock Autoaugment: Learning augmentation policies from data.
\newblock \emph{arXiv preprint arXiv:1805.09501}, 2018.

\bibitem[DeVries \& Taylor(2017)DeVries and Taylor]{devries2017cutout}
Terrance DeVries and Graham~W Taylor.
\newblock Improved regularization of convolutional neural networks with cutout.
\newblock \emph{arXiv preprint arXiv:1708.04552}, 2017.

\bibitem[He et~al.(2016)He, Zhang, Ren, and Sun]{he2016deep}
Kaiming He, Xiangyu Zhang, Shaoqing Ren, and Jian Sun.
\newblock Deep residual learning for image recognition.
\newblock In \emph{Proceedings of the IEEE conference on computer vision and
  pattern recognition}, pp.\  770--778, 2016.

\bibitem[Howard et~al.(2017)Howard, Zhu, Chen, Kalenichenko, Wang, Weyand,
  Andreetto, and Adam]{howard2017mobilenets}
Andrew~G Howard, Menglong Zhu, Bo~Chen, Dmitry Kalenichenko, Weijun Wang,
  Tobias Weyand, Marco Andreetto, and Hartwig Adam.
\newblock Mobilenets: Efficient convolutional neural networks for mobile vision
  applications.
\newblock \emph{arXiv preprint arXiv:1704.04861}, 2017.

\bibitem[Hu et~al.(2018)Hu, Shen, and Sun]{hu2018squeeze}
Jie Hu, Li~Shen, and Gang Sun.
\newblock Squeeze-and-excitation networks.
\newblock In \emph{Proceedings of the IEEE conference on computer vision and
  pattern recognition}, pp.\  7132--7141, 2018.

\bibitem[Krizhevsky et~al.(2012)Krizhevsky, Sutskever, and
  Hinton]{krizhevsky2012imagenet}
Alex Krizhevsky, Ilya Sutskever, and Geoffrey~E Hinton.
\newblock Imagenet classification with deep convolutional neural networks.
\newblock \emph{Advances in neural information processing systems},
  25:\penalty0 1097--1105, 2012.

\bibitem[Kuang(2021)]{liukuang2021}
Liu Kuang.
\newblock Liu/pytorch-cifar.
\newblock \url{https://github.com/kuangliu/pytorch-cifar}, 2021.

\bibitem[{Le~Digabel}(2011)]{Le09b}
S.~{Le~Digabel}.
\newblock {Algorithm 909: NOMAD: Nonlinear Optimization with the MADS
  algorithm}.
\newblock \emph{{ACM} Transactions on Mathematical Software}, 37\penalty0
  (4):\penalty0 44:1--44:15, 2011.
\newblock \doi{10.1145/1916461.1916468}.
\newblock URL \url{http://dx.doi.org/10.1145/1916461.1916468}.

\bibitem[Simonyan \& Zisserman(2014)Simonyan and Zisserman]{simonyan2014very}
Karen Simonyan and Andrew Zisserman.
\newblock Very deep convolutional networks for large-scale image recognition.
\newblock \emph{arXiv preprint arXiv:1409.1556}, 2014.

\bibitem[Szegedy et~al.(2015)Szegedy, Liu, Jia, Sermanet, Reed, Anguelov,
  Erhan, Vanhoucke, and Rabinovich]{szegedy2015going}
Christian Szegedy, Wei Liu, Yangqing Jia, Pierre Sermanet, Scott Reed, Dragomir
  Anguelov, Dumitru Erhan, Vincent Vanhoucke, and Andrew Rabinovich.
\newblock Going deeper with convolutions.
\newblock In \emph{Proceedings of the IEEE conference on computer vision and
  pattern recognition}, pp.\  1--9, 2015.

\bibitem[Szegedy et~al.(2016)Szegedy, Vanhoucke, Ioffe, Shlens, and
  Wojna]{szegedy2016rethinking}
Christian Szegedy, Vincent Vanhoucke, Sergey Ioffe, Jon Shlens, and Zbigniew
  Wojna.
\newblock Rethinking the inception architecture for computer vision.
\newblock In \emph{Proceedings of the IEEE conference on computer vision and
  pattern recognition}, pp.\  2818--2826, 2016.

\bibitem[Szegedy et~al.(2017)Szegedy, Ioffe, Vanhoucke, and
  Alemi]{szegedy2017inception}
Christian Szegedy, Sergey Ioffe, Vincent Vanhoucke, and Alexander Alemi.
\newblock Inception-v4, inception-resnet and the impact of residual connections
  on learning.
\newblock In \emph{Proceedings of the AAAI Conference on Artificial
  Intelligence}, volume~31, 2017.

\bibitem[Tan \& Le(2019)Tan and Le]{tan2019efficientnet}
Mingxing Tan and Quoc Le.
\newblock Efficientnet: Rethinking model scaling for convolutional neural
  networks.
\newblock In \emph{International Conference on Machine Learning}, pp.\
  6105--6114. PMLR, 2019.

\bibitem[Taylor \& Nitschke(2017)Taylor and Nitschke]{taylor2017improving}
Luke Taylor and Geoff Nitschke.
\newblock Improving deep learning using generic data augmentation.
\newblock \emph{arXiv preprint arXiv:1708.06020}, 2017.

\bibitem[Vinyals et~al.(2016)Vinyals, Blundell, Lillicrap, Kavukcuoglu, and
  Wierstra]{vinyals2016matching}
Oriol Vinyals, Charles Blundell, Timothy Lillicrap, Koray Kavukcuoglu, and Daan
  Wierstra.
\newblock {Matching Networks for One Shot Learning}.
\newblock In \emph{Proceedings of the 30th International Conference on Neural
  Information Processing Systems}, pp.\  3637--3645, 2016.

\end{thebibliography}
